# Optimistic and Pessimistic Neural Networks for Scene and Object Recognition

Rene Grzeszick   Sebastian Sudholt   Gernot A. Fink

TU Dortmund University

{rene.grzeszick,sebastian.sudholt,gernot.fink}@tu-dortmund.de

## Abstract

*In this paper the application of uncertainty modeling to convolutional neural networks is evaluated. A novel method for adjusting the network's predictions based on uncertainty information is introduced. This allows the network to be either optimistic or pessimistic in its prediction scores. The proposed method builds on the idea of applying dropout at test time and sampling a predictive mean and variance from the network's output. Besides the methodological aspects, implementation details allowing for a fast evaluation are presented. Furthermore, a multilabel network architecture is introduced that strongly benefits from the presented approach. In the evaluation it will be shown that modeling uncertainty allows for improving the performance of a given model purely at test time without any further training steps. The evaluation considers several applications in the field of computer vision, including object classification and detection as well as scene attribute recognition.*

## 1. Introduction

Convolutional Neural Networks (CNNs) show state-of-the-art results in many computer vision applications. Their usage includes scene [33] and object classification [14, 28], object detection [25, 9], scene parsing [6], face recognition [20], medical imaging [26] and many more. Lately, their applications even extended to non-vision tasks like the recognition of acoustic scenes [22] or playing games [3].

While CNNs work well in practice, it has been shown that they are typically over-confident in their predictions [17]. It is possible to generate arbitrary images that are nevertheless classified with high scores after the softmax computation. There is no measure of uncertainty associated with their output. Compared to many traditional pattern recognition approaches this is a major shortcoming [2, 4]. The most prominent example are Bayesian models [2]. A typical approach is, for example, model averaging where predictions are made by a set of plausible models and their predictions are integrated into a single representative one [2, 19]. This allows for computing uncertainty based on a predictive mean and variance. Uncertainty is often similarly modeled in regression tasks. Bayesian approaches and also Support Vector Regressors allow for computing confidence bands based on a variance estimation [4, 23], which results in an interval around the regressed values.

The issue of uncertainty modeling for Deep Neural Networks has recently been addressed in [8, 11]. In [11] it is argued that dropout models the training of a set of different network models. Dropout is a mechanism that is frequently used at training time in order to avoid overfitting [30]. A random set of neurons is dropped from the network so that multiple paths through the network are learned. This can also be thought of as learning an ensemble of different classifiers that are based on different combinations of neurons. The number of classifiers in the ensemble is exponential in the number of neurons and all networks make use of heavy parameter sharing [30]. In [11] it is argued that a forward pass through the network in which the weights are divided by two is an approximation of model averaging for multilayer perceptrons. This approach is then combined with maxout neurons. A maxout neuron simply returns the maximum activation from a set of inputs. A maxout network learns to have roughly the same output regardless which inputs are dropped and therefore interacts well with dropout.

In [8] a similar approach has been introduced. A connection has been drawn between Deep Neural Networks and Gaussian processes [24]. Without changing the neural network, it can be shown that applying dropout to each layer can theoretically be interpreted as a Bayesian approximation of Gaussian Processes. As a result, a predictive mean and variance can be computed from the results of multiple forward passes through the network where different neurons are dropped which then allows for associating uncertainty with the network's output. The computed predictive mean is basically another approach to model averaging [2] and improves the performance compared to the implicit model averaging that is performed by computing a single pass through the network.

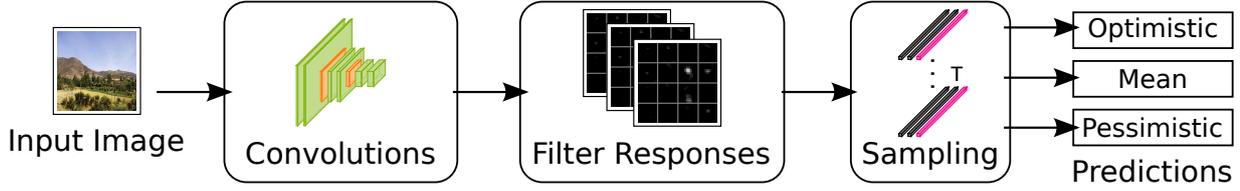

Figure 1. Overview of the proposed architecture. The convolution layers of the CNN are evaluated in a single forward pass, computing the input features for the fully connected part. Multiple passes through the fully connected part using dropout are then used for computing a robust prediction.

The contribution of this paper is an extension of the theoretical model provided in [8]. Based on this theoretical model, optimistic or pessimistic network behavior is defined. Implementation details, providing a fast evaluation of neural networks under the given model are shown. An evaluation on various computer vision tasks, including object classification and detection, as well as the prediction of multiple object classes and scene attributes, is provided.

## 2. Review of Test Dropout

In the following section the idea presented in [8] is briefly reviewed. It has been shown that a deep neural network with dropout applied before every weight layer is mathematically equivalent to an approximation of a deep Gaussian process. This model allows for the computation of a predictive mean and variance.

Let $W_i^j$ be a binary variable for the $i^{th}$ node in layer $j$. Each binary variable takes the value 1 with probability $p_j$ for the networks layer $j$. It is dropped out if the binary variable is set to 0. Given $T$ repetitions and $J$ layers a random set of variables $\omega_t = \{W\}_i^J$ is randomly drawn for each repetition $t$. For the input value $x$ and the prediction $\hat{y}_t(x, \omega_t)$ at repetition $t$, the predictive mean

$$E(y) \approx \tau^{-1} \boldsymbol{I}_C + \frac{1}{T} \sum_{t=1}^{T} \hat{y}_t(x, \omega_t) \qquad (1)$$

and predictive variance

$$Var(y) \approx \tau^{-1} \boldsymbol{I}_C +$$
$$\frac{1}{T} \sum_{t=1}^{T} \hat{y}_t(x, \omega_t)^T \hat{y}_t(x, \omega_t) - E(y)^T E(y) \qquad (2)$$

can be computed with $\boldsymbol{I}_C$ denoting a vector of ones of length $C$ which equals the number of classes at the output layer of the network. Furthermore, the standard deviation is denoted by $\sigma(y) = \sqrt{Var(y)}$. In this formulation $\tau$ refers to the model precision:

$$\tau = \frac{pl^2}{2N\lambda} \qquad (3)$$

with $\lambda$ being the weight-decay parameter, the prior length scale $l$, $N$ the number of the input samples and $p$ being the inverse of the dropout probability of the network (c.f. [8]). This reasoning can be applied to every neural network using standard dropout.

## 3. Optimistic and Pessimistic Networks

Following this theoretical model, a novel modeling of network behavior is proposed. Using confidence intervals, a neural network can either be optimistic or pessimistic in its predictions. The predictive mean and both network behaviors are then applied to computer vision tasks.

### 3.1. Confidence Intervals

Under the assumption that the output of a deep Gaussian process and therefore the neural network is normal distributed, it is possible to compute a confidence interval for the predictions based on the mean and standard deviation:

$$[E(y) - z_{1-\frac{\alpha}{2}} \frac{\sigma(y)}{\sqrt{T}} \;;\; E(y) + z_{1-\frac{\alpha}{2}} \frac{\sigma(y)}{\sqrt{T}}] \qquad (4)$$

with $z_{1-\frac{\alpha}{2}}$ being the $1 - \alpha/2$ quantile of the normal distribution. Choosing an appropriate value for $\alpha$, a confidence interval can be derived (i.e. for 99% certainty, $\alpha = 0.01$).

Note that given the mean and variance, i.e. from a validation set, a guaranteed confidence interval can be computed based on the number of runs $T$:

$$T = \left( z_{(1-\frac{\alpha}{2})} \frac{\sigma(y)}{E(y)} \right)^2 \qquad (5)$$

Adjusting the parameter $T$ it can be said that the performance of the model will be within the respective confidence interval with a probability of $\alpha$.

### 3.2. Definition of Optimistic and Pessimistic Networks Behavior

The result that is obtained from computing the confidence interval provides an upper and lower bound for the prediction which is dependent on the confidence level. Hence, the true mean is between these bounds. The goal is to use this information in order to improve the results of a convolutional neural network.

While using dropout and computing the predictive mean should already be more robust than a single pass through the network (cf. [7]), it does not consider the variance that is obtained from scoring the sample multiple times. Given two samples $x_1$ and $x_2$ with the same mean but different confidence intervals it can be argued that one should be scored better than the other. This gives rise to optimistic or pessimistic network behavior, which are defined by

$$E_o(y) = E(y) + z_{1-\frac{\alpha}{2}} \frac{\sigma(y)}{\sqrt{T}} \qquad (6)$$

and

$$E_p(y) = E(y) - z_{1-\frac{\alpha}{2}} \frac{\sigma(y)}{\sqrt{T}} \qquad (7)$$

respectively. This can be interpreted by the reasoning that dropout models an ensemble which is generated by multiple paths through the network [30]. The two behaviors can be compared to a decision rule of ensemble classifiers such as the maximum rule (c.f. [15]). The optimistic behavior gives more weight to the numerically larger output of the network. This is essentially a bias toward a minority of votes in the ensemble determining neurons in the output layer to be active. The pessimistic formulation favors a minority of votes which are in favor of a neuron in the output layer being inactive and thus to numerically smaller values. In the experiments it will be shown that modifying the network to output a more optimistic or more pessimistic score can improve the results.

Note that for a classification network that has been trained on a given set of images, $\tau$ is not only constant during test time, but in most practical cases the model uncertainty can be considered small compared to the empirical mean and variance. For example, ImageNet has more than 1.2 million images and the number of samples increases even more when considering data augmentation. Using typical parameters for the dropout $p = 0.5$, the length scale $l^2 = 0.005$ and the weight decay $\lambda = 0.0001$ (cf. [14, 8]), it follows that the model precision $\tau \approx 0.0000104$. Hence, both, $E(y)$ and $\sigma(y)$, can be approximated based on the empirical mean and variance.

### 3.3. Fast Optimistic and Pessimistic Networks

Theoretically, processing $T$ forward passes would come at a $T$ times higher computation time. Although in practice parallelization using a larger batch size (if possible $T$) allows for a slightly faster processing, the increased computational cost makes this approach unsuitable for many practical applications.

In typical CNN architectures dropout is only applied during training and most classification networks restrict themselves to dropout in the last weight layers in order to avoid overfitting [14, 28]. Modeling this for the sampling at test time can be used as an advantage. Following the idea of [25] it is assumed that the network consists of two parts: The first part is a fixed set of convolution layers. The activations of these layers are considered as an intermediate feature representation and fixed for a given input image. The second part is the neural network using dropout. The resulting architecture is illustrated in Fig. 1. The convolutions are computed in a single forward pass, then $T$ passes through the fully connected part are computed. The dropout value for each of the $j$ layers of the deep neural network is set to a single value $p_{drop}$.

## 4. Evaluation

The goal of the evaluation is to show the performance improvement that is achieved by evaluating the same models using dropout at test time. A single forward pass without dropout is compared with multiple passes using dropout and computing a predictive mean or the output of the proposed optimistic or pessimistic network behavior. No additional training step is performed. Different tasks have been evaluated. Namely, the ILSVRC 2014 object classification task, the VOC2011 object detection task using R-CNNs and the SUN scene attributes as well as the VOC2011 object prediction task using a multilabel network have been evaluated.

### 4.1. Network Architectures

The behavior is investigated for different network architectures based on the given task. All architectures are sketched in Tab. 1. In principle all of them are similar to the VGG16 network architecture. In its original design the VGG16 architecture uses a convolution part with 13 layers of $3 \times 3$ convolutions in combination with pooling operations. The convolution part is followed by a fully connected part with three layers and a softmax classifier at the end [28]. The VGG16 architecture is evaluated on the ILSVRC classification task [28] and integrated in the R-CNN model [9] as well.

In addition to the original VGG16 architecture, two networks for multilabel prediction have been evaluated. The architectures are similar to the ones presented in [1, 31]. The multilabel networks are based on the VGG16 architecture, but instead of predicting one label with a softmax layer, a sigmoid layer is used in order to derive pseudo-probabilities for multiple labels. This approach can be used for predicting the presence of multiple objects or attributes of scenes or objects in an integrated manner and is known as *logistic regression*.

For training the network, a loss function has to be defined which is suitable for the regression task. Historically, the loss function of choice has been the Euclidean loss

$$E_{L2} = \frac{||\mathbf{y} - \hat{\mathbf{y}}||_2^2}{2} \qquad (8)$$

| VGG16 | VGG16 multilabel | VGG16 fully conv |
|---|---|---|
| input 224 × 224 | variable input | |
| conv3-64 | | |
| conv3-64 | | |
| maxpool | | |
| conv3-128 | | |
| conv3-128 | | |
| maxpool | | |
| conv3-256 | | |
| conv3-256 | | |
| conv3-256 | | |
| maxpool | | |
| conv3-512 | | |
| conv3-512 | | |
| conv3-512 | | |
| maxpool | | |
| conv3-512 | | |
| conv3-512 | | |
| conv3-512 | | |
| maxpool | SPP | maxpool |
| FC-4096 * | FC-4096 * | conv3-512 * |
| FC-4096 * | Maxout-2048 | conv3-#classes |
| FC-#classes | FC-#classes | global max |
| Softmax | Sigmoid | Sigmoid |

Table 1. Overview of the network architectures. (left) VGG16 as introduced in [28]. (right) The proposed multiclass architecture. Dropout is applied to the layers marked with an asterisk.

Where $\mathbf{y}$ and $\hat{\mathbf{y}}$ are the label vector and prediction vector of the CNN respectively. However, using the Euclidean loss in combination with a logistic regression can lead to slow training as the gradient of the combined sigmoid activation and loss is scaled by the derivative of the sigmoid. If the CNN produces large absolute values, this leads to vanishing gradients. In order to eliminate this problem, the *Cross Entropy loss*

$$E_{CE} = -\sum_{i=1}^{d} \mathbf{y_i} \log \hat{\mathbf{y}}_\mathbf{i} + (1 - \mathbf{y_i}) \log(1 - \hat{\mathbf{y}}_\mathbf{i}) \quad (9)$$

is used for logistic regression tasks, e.g. in [12, 32]. Using this loss, the combined gradient is no longer scaled by the derivative of the sigmoid activation. The same loss function is adapted for the experiments using the multilabel network architectures.

The first multilabel architecture is very similar to the original VGG16 design. As already mentioned above, the softmax layer is replaced by a sigmoid layer. Additionally, the second fully connected layer is replaced by a maxout layer [11] and the last pooling layer is replaced with a spatial pyramid pooling layer [13]. This way, the CNN is able to process images of different sizes without the need for rescaling or cropping. The second architecture is a fully convolutional network (c.f. [29]). This architecture is applied to the task of object prediction. Here, it is assumed that each object is visually distinguishable by the filters and therefore recognizable by a fully convolutional network in a single forward pass. The fully connected layers of the VGG16 architecture are replaced by two additional convolution layers: the first layer has 512 filters and the second layer has one filter per target class. Using a global max pooling for indicating the object presence, the last layer basically replaces the class scores which are derived from the last fully connected layer in the VGG16 architecture.

In all network architectures the first thirteen weight layers are considered as the feature representation and dropout is then applied to the remainder of the network. The respective layers are marked with an asterisk in Tab. 1. For the networks using a fully connected structure, these are the first two fully connected layers so that the third layer models the scores for each of the classes. For the fully convolutional architecture, dropout is applied to the first additional convolution layer so that the global max pooling from the final layer represents the class scores.

### 4.2. Significance Testing

In order to assess the significance of the differences in performance a statistical test, namely a permutation test [10, 18] (also known as randomization test) is evaluated on the results of the classification task. For the retrieval tasks, running a statistical test is impossible as only the baseline's mean average precision is available. In order to run a statistical test, the average precisions would be needed. This is different from the classification task, where a classification is either correct or not. Here, a statistic can be reconstructed simply from the number of test samples and the test performance.

The permutation test is chosen over other ones such as the ubiquitous t-test as the permutation test makes no assumptions about the distribution of the data. An exact permutation test is not feasible for the ILSVRC 2014 as all permutations are impossible to compute due to the test set size of 50 000. Instead, the exact statistic is approximated by a finite amount of $n$ permutations with

$$n = \frac{p(1-p)}{\sigma_p^2}. \quad (10)$$

Here, $p$ is the $p$-value and $\sigma_p$ is the desired upper bound of the standard deviation of $p$ induced by the approximate statistic [10]. Since $p$ is unknown beforehand, the upper bound for $n$ can be approximated by setting $p = 0.5$. Furthermore, setting the desired $\sigma_p$ to 0.001, 250 000 permuta-tions have to be computed in order to assess the significance of the results.

| Network      | $T$ | $p_{drop}$ | top-1 error | top-3 error | top-5 error |
|--------------|-----|------------|-------------|-------------|-------------|
| VGG16 (*) [28] | -   | -          | 28.6%       | 13.9%       | 10.1%       |
| Mean         | 10  | 0.1        | **27.1%**   | **12.5%**   | **8.8%**    |
| Optimistic   | 10  | 0.1        | 27.3%       | 12.6%       | 8.9%        |
| Pessimistic  | 10  | 0.1        | 27.2%       | 12.9%       | 9.4%        |
| Mean         | 100 | 0.1        | 27.1%       | 12.5%       | 8.9%        |
| Optimistic   | 100 | 0.1        | 27.1%       | 12.6%       | 8.9%        |
| Pessimistic  | 100 | 0.1        | 27.1%       | 12.5%       | 8.9%        |
| Mean         | 10  | 0.5        | 27.2%       | 12.8%       | 9.0%        |
| Optimistic   | 10  | 0.5        | 27.3%       | 13.1%       | 9.3%        |
| Pessimistic  | 10  | 0.5        | 27.6%       | 17.7%       | 16.5%       |
| Mean         | 100 | 0.5        | 26.9%       | **12.4%**   | **8.8%**    |
| Optimistic   | 100 | 0.5        | 26.9%       | 12.5%       | 8.9%        |
| Pessimistic  | 100 | 0.5        | **26.8%**   | **12.4%**   | **8.8%**    |

Table 2. Error rates for different values of $T$ and $p_{drop}$ on the ILSVRC 2014 classification on the validation set. (*) Results were re-evaluated using the publicly available model.

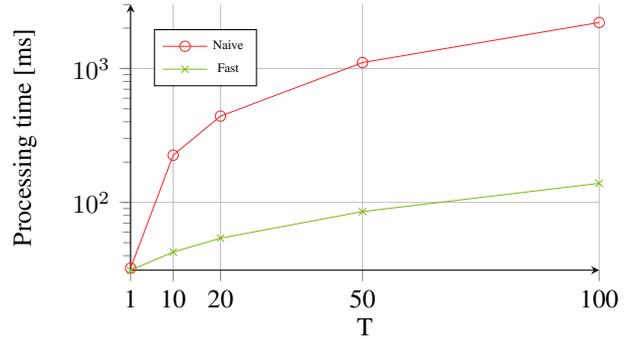

Figure 2. Computation time in ms on the ILSVRC benchmark on a TitanX using dense sampling with mirroring and a scale size of 256. The runtime in relation to the sampling parameter $T$ is evaluated.

### 4.3. ILSVRC 2014

The VGG16 network has been evaluated on the validation set of the ILSVRC-2014 classification task [27]. In [28] multiple crops which are extracted at different image scales are evaluated by the CNN. Moreover, the crops are larger than the CNNs input size and then densely evaluated by reshaping the fully connected layers to $7 \times 7$ and $1 \times 1$ convolution layers. The authors stated that extracting multiple crops at different image scales only slightly improves the performance at the cost of a much higher computation time. This evaluation is therefore omitted. Evaluating 150 different crops in [28] improved the results by $1.3\%$. Still the dense evaluation at a single scale of 256 (shortest side length) is performed. The publicly available model from [28] is used for the evaluation.

The network has then been evaluated using the mean, optimistic and pessimistic behavior with different parameters for $p_{drop}$ and $T$. The maximum activation after the softmax computation from each of the $T$ dense samplings is used. Note that averaging over the dense sampling would not be beneficial as changes caused by the dropout would be reduced when averaging the results. The error rates for the top $k$ results are shown in Fig. 2. The top-1 error rate can be improved by $1.8\%$ and the top-5 error rate can be improved by $1.3\%$ compared to a single forward pass. However, there is not much difference between the different network behaviors for this task.

The permutation tests for top-1, top-3 and top-5 error all showed a significant difference between the pessimistic net's results and the baseline results with a $p$-value of $p < 0.01$ each.

Not surprisingly the results converge when increasing the number of runs $T$. Note that with a higher dropout value $p_{drop}$ the number of runs $T$ that are required to improve the result increases. This can be explained by the fact that the variance will be increased the more information is dropped from the model. Overall the results are improved when using a higher dropout ratio. A crucial point is that this improved performance comes at the cost of a higher runtime. Therefore, the computation time for $T$ passes through the network have been evaluated with the results shown in Fig. 2. The fast architecture is compared with a naive approach computing $T$ forward passes through the complete network. While the naive approach is almost linear in time, the computation time can be reduced by a large margin when evaluating the fast implementation that pre-computes the filter responses.

### 4.4. VOC2011 - Detection

Besides the pure classification, an object detection task has been evaluated on the VOC2011 dataset [5]. Therefore, the faster R-CNN has been evaluated [25, 9]. It uses a region proposal network in order to generate region candidates that might contain an object. These region candidates are then evaluated by a CNN using the VGG16 architecture. The model as provided in [9] is used which is also pre-trained on ImageNet and then adapted to the object detection task on the VOC dataset. The faster R-CNN architecture is already designed to pre-compute the filter responses and then evaluate the region proposals as well as the classification based on those filters. Hence, the proposed sampling is simply applied to the object classification part of the R-CNN. As for the ILSVRC experiment, a softmax classifier is used for predicting the class or assigning the region candidate to the background.

| Network | T | $p_{drop}$ | 300 Regions | 600 Regions |
|---|---|---|---|---|
| R-CNN (*) [9] | - | - | 64.7% | 64.5% |
| Optimistic | 10 | 0.1 | **65.9%** | **65.9%** |
| Mean | 10 | 0.1 | 65.7% | 65.8% |
| Pessimistic | 10 | 0.1 | 65.8% | 66.0% |
| Optimistic | 100 | 0.5 | 65.7% | **65.8%** |
| Mean | 100 | 0.5 | **65.8%** | 65.7% |
| Pessimistic | 100 | 0.5 | 65.7% | 65.4% |

Table 3. mAP for R-CNNs on the VOC 2011 dataset. (*) Results were re-evaluated using the publicly available model.

The evaluation computes the mean average precision (mAP) over all classes. An intersection over union (IoU) of 50% between a prediction and a ground truth bounding box is required in order to count a prediction as correct [5]. The results are shown in Tab. 3. Similarly to the ILSVRC results which are also based on a softmax classification, there is an improvement in the detection rates but there is not much difference between the different network behaviors. The mAP can be improved by about 1.2% compared to a single pass through the network. In contrast to the ILSVRC task a dropout rate $p_{drop} = 0.1$ is already sufficient to improve the results and increasing the dropout together with the number of sampling runs does not improve the detection rates.

### 4.5. SUN attributes

The SUN-Attributes dataset focuses on scene understanding [21]. Each of the scene images has been annotated with additional attributes such as open, enclosed, natural or manmade. In total there are 102 different attributes. Each attribute is annotated as present or absent by three different annotators. Following the official evaluation protocol from [21], 90% of the dataset are used for training and 10% for testing. Furthermore, only high confidence annotations are considered. An attribute is considered as present if it got at least two votes and it is considered as absent if it got zero votes. The recognition of scene attributes is of interest as predicting these attributes allows for zero-shot recognition. In this scenario unknown scene categories are recognized by describing them in terms of their attributes [16, 21]. Therefore, an accurate attribute prediction is important.

As a scene can be associated with multiple attributes this is also a perfect application for the presented multilabel architecture. The network's convolution layers are pre-trained on ImageNet as this typically improves the results and the attribute prediction is then trained on the SUN-Attributes training set. 500 000 samples have been created using data augmentation, including Gaussian noise ($\sigma = 0.02$), random translations (up to 5%) and rotations (up to 10°). A batch size of 256 has been used for training with 4 000 iterations using an initial learning rate of 0.0005. The learning rate has then reduced by the factor of 10 and a final 2 000 training iterations have been computed. When training the network with the cross entropy loss, it implicitly learns the co-occurrences of different classes, i.e. the scene attributes.

The results for the attribute recognition are shown in Tab. 4. Traditional approaches for attribute recognition evaluate one SVM per attribute [21]. Note that the results of the VGG16 multilabel architecture already improve the results compared to the traditional SVM approach. Applying the dropout to the evaluations allows for a slight improvement of the results. It seems to be easier to improve the results of some of the categories. For example, the affordances and spatial envelope attributes. Note that there is a high ambiguity in the labels of the SUN dataset which may make it difficult to further improve on the results. Overall, there is a slight tendency toward an optimistic evaluation.

### 4.6. VOC2011 - Multiclass prediction

In the last experiments, the presence of multiple objects in a scene is predicted. Therefore, the fully convolutional VGG16 multilabel architecture has been evaluated on the VOC2011 dataset [5]. The object presence is predicted simultaneously based on the pseudo-probabilities of the per-class sigmoid computation. The original VGG16 network is not able to predict the presence of multiple objects, instead for each object class an SVM is trained on the 4096 dimensional feature vector from the fc7-layer [28].

The filter functions are pre-trained on ImageNet and the presence prediction is then trained on the VOC2011 training set. 500 000 samples have been created using data augmentation, again including Gaussian noise, random translations and rotations (using the same parameters as for the SUN-Attributes dataset). A batch size of 256 has been used for training with 2000 iterations using a learning rate of 0.001. The learning rate has then been reduced to 0.0001 computing 2 000 more training iterations.

The images from the VOC2011 dataset have been rescaled so that the shortest side has a length of 512px. Note that this is a rather large image size compared to many other tasks [14, 28]. As the network is fully convolutional this image size is beneficial since the objects are larger in terms of the absolute surface area and can be detected more easily by the large receptive field of the network's filters.

The mAP over all classes is computed and the results are shown in Tab. 5. As the predictions are evaluated independent of each other (i.e. two classes can be predicted with a probability of 1), the proposed behavior modeling has more influence than in case of a softmax classifier. Also, the mAP is evaluated based on varying thresholds for each of the 20 classes in the VOC dataset. As a result, the influence of the different network behaviors is clearly visible. The mAP can be improved by up to 1% when using an optimistic behavior instead of the predictive mean and by 2.5% compared to the plain evaluation.

|         |    |            | mAP |  |  |  |
| Network | $T$ | $p_{drop}$ | Affordances | Materials | Surfaces | Spatial Envelope |
| --- | --- | --- | --- | --- | --- | --- |
| per class SVMs [21] | - | - | 44% | 51% | 50% | 62% |
| VGG16 multilabel | - | - | 54.2% | 58.6% | 58.5% | 65.2% |
| Optimistic | 10 | 0.1 | **54.3%** | **58.9%** | 58.4% | **65.4%** |
| Mean | 10 | 0.1 | **54.3%** | 58.7% | 58.3% | 65.3% |
| Pessimistic | 10 | 0.1 | 54.1% | 58.7% | 58.2% | 65.2% |
| Optimistic | 100 | 0.5 | **54.7%** | 58.5% | 58.7% | **65.5%** |
| Mean | 100 | 0.5 | **54.7%** | 58.4% | **58.8%** | 65.4% |
| Pessimistic | 100 | 0.5 | 54.6% | 58.3% | 58.6% | 65.4% |

Table 4. Mean average precision for the attribute prediction on the SUN attributes dataset.

| Network | $T$ | $p_{drop}$ | mAP |
| --- | --- | --- | --- |
| Multiclass baseline | - | - | 72.8% |
| Optimistic | 10 | 0.1 | **74.6%** |
| Mean | 10 | 0.1 | 73.5% |
| Pessimistic | 10 | 0.1 | 73.2% |
| Optimistic | 100 | 0.5 | **75.3%** |
| Mean | 100 | 0.5 | 74.8% |
| Pessimistic | 100 | 0.5 | 74.6% |

Table 5. Mean average precision for the presence prediction on the VOC2011 dataset.

### 4.7. Discussion

All of the proposed networks use some kind of maximum pooling. The ILSVRC network pools the maximum response of the dense sampling over the image, the R-CNN uses a non-maximum suppression of the bounding box detections and both multilabel architectures have a pooling layer that is applied after the layers that are dropped out. It is worth noting that these are important design decision when applying dropout. This also confirms the findings in [11] where it is argued that maxout layers and dropout in training interact well. A reason is that when applying average pooling operations the network smooths the differences that are generated by choosing different paths through the network.

The strongest influence can be observed on the ILSVRC task and the VOC object prediction task, where the maxpooling is applied in the last layer. In contrast, the lowest influence is observed on the scene attributes task, where the last layer is a fully connected layer that follows the maxout output. It seems that the last network layer learns to produce roughly the same output even when neurons are dropped.

Furthermore, it is interesting that the difference between a predictive mean and a more optimistic or pessimistic estimation of the networks prediction is more clearly visible when predicting scores for many values at once.

### 5. Conclusion

In this paper the application of uncertainty modeling to convolutional neural networks has been evaluated. The proposed method builds on the idea of applying dropout at test time and sampling a predictive mean and variance from the network's output which in turn also allows for for adjusting the networks predictions to be more optimistic or more pessimistic. In general the proposed method can be compared to evaluating an ensemble classifier within a single network architecture with the different behaviors representing different decision rules.

Besides the methodological framework, implementation details are provided. An approach allowing for an efficient evaluation at test time and network architectures that solve multilabel tasks in an integrated manner and interact well with the proposed approach are presented.

It could be shown that the results of a given model can be improved on scene and object recognition tasks. For the scene attributes task, the proposed multilabel architecture is able to outperform the traditional approach using per class SVMs by far and the results are further improved using test dropout. The evaluation also showed that the proposed scheme interacts well with methods that build on maxpooling, exploiting the differences that are obtained by computing several maxima instead of averaging. The most prominent example being the results on the VOC dataset where the highest responses from a fully convolutional network are used in order to predict the presence of multiple objects. Here, the results can be improved by as much as 2.5% using an optimistic evaluation scheme and dropout at test time.

### Acknowledgement

This work has been supported by the German Research Foundation (DFG) within project Fi799/9-1 ('Partially Supervised Learning of Models for Visual Scene Recognition').

# References

[1] S. Bach, A. Binder, G. Montavon, K.-R. Müller, and W. Samek. Analyzing classifiers: Fisher vectors and deep neural networks. *arXiv preprint arXiv:1512.00172*, 2015.

[2] D. Barber. *Bayesian reasoning and machine learning*. Cambridge University Press, 2012.

[3] C. Clark and A. Storkey. Teaching deep convolutional neural networks to play go. *arXiv preprint arXiv:1412.3409*, 2014.

[4] K. De Brabanter, P. Karsmakers, J. De Brabanter, J. A. Suykens, and B. De Moor. Confidence bands for least squares support vector machine classifiers: A regression approach. *Pattern Recognition*, 45(6):2280–2287, 2012.

[5] M. Everingham, L. Van Gool, C. K. I. Williams, J. Winn, and A. Zisserman. The PASCAL Visual Object Classes Challenge 2011 (VOC2011) Results. http://www.pascal-network.org/challenges/VOC/voc2011/workshop/index.html, 2011.

[6] C. Farabet, C. Couprie, L. Najman, and Y. LeCun. Learning hierarchical features for scene labeling. *IEEE Transactions on Pattern Analysis and Machine Intelligence*, 35(8):1915–1929, 2013.

[7] Y. Gal and Z. Ghahramani. Dropout as a Bayesian approximation: Insights and applications. In *Deep Learning Workshop, Int. Conf. on Machine Learning (ICML)*, 2015.

[8] Y. Gal and Z. Ghahramani. Dropout as a Bayesian Approximation: Representing Model Uncertainty in Deep Learning. In *Proc. Int. Conf. on Machine Learning (ICML)*, 2016.

[9] R. Girshick, J. Donahue, T. Darrell, and J. Malik. Region-based convolutional networks for accurate object detection and segmentation. *IEEE Transactions on Pattern Analysis and Machine Intelligence*, 38(1):142–158, 2016.

[10] P. Good. *Permutation Tests - A Practical Guide to Resampling Methods for Testing Hypothesis*. Springer, 2 edition, 2000.

[11] I. J. Goodfellow, D. Warde-Farley, M. Mirza, A. C. Courville, and Y. Bengio. Maxout Networks. *ICML (3)*, 28:1319–1327, 2013.

[12] E. M. Hand and R. Chellappa. Attributes for Improved Attributes: A Multi-Task Network for Attribute Classification. *arXiv*, 2016.

[13] K. He, X. Zhang, S. Ren, and J. Sun. Spatial Pyramid Pooling in Deep Convolutional Networks for Visual Recognition. *IEEE transactions on pattern analysis and machine intelligence*, 37(9):1904–1916, 2015.

[14] A. Krizhevsky, I. Sutskever, and G. E. Hinton. Imagenet classification with deep convolutional neural networks. In *Advances in neural information processing systems*, pages 1097–1105, 2012.

[15] L. Kuncheva. Combining Pattern Classifiers: Methods and Algorithms, 2004.

[16] C. H. Lampert, H. Nickisch, and S. Harmeling. Attribute-based classification for zero-shot visual object categorization. *IEEE Transactions on Pattern Analysis and Machine Intelligence*, 36(3):453–465, 2014.

[17] A. Nguyen, J. Yosinski, and J. Clune. Deep neural networks are easily fooled: High confidence predictions for unrecognizable images. In *Proc. IEEE Conf. on Computer Vision and Pattern Recognition (CVPR)*, pages 427–436. IEEE, 2015.

[18] M. Ojala and G. C. Garriga. Permutation Tests for Studying Classifier Performance. *Journal of Machine Learning Research*, 11:1833–1863, 2010.

[19] I. Park, H. K. Amarchinta, and R. V. Grandhi. A bayesian approach for quantification of model uncertainty. *Reliability Engineering & System Safety*, 95(7):777–785, 2010.

[20] O. M. Parkhi, A. Vedaldi, and A. Zisserman. Deep face recognition. In *British Machine Vision Conference*, 2015.

[21] G. Patterson, C. Xu, H. Su, and J. Hays. The sun attribute database: Beyond categories for deeper scene understanding. *International Journal of Computer Vision*, 108(1-2):59–81, 2014.

[22] H. Phan, L. Hertel, M. Maass, P. Koch, and A. Mertins. CNN-LTE: a Class of 1-X Pooling Convolutional Neural Networks on Label Tree Embeddings for Audio Scene Recognition. *arXiv preprint arXiv:1607.02303*, 2016.

[23] C. M. B. C. S. Qazaz. Regression with input-dependent noise: A bayesian treatment. In *Advances in Neural Information Processing Systems 9: Proceedings of the 1996 Conference*, volume 9, page 347. MIT Press, 1997.

[24] C. E. Rasmussen. Gaussian processes for machine learning. 2006.

[25] S. Ren, K. He, R. Girshick, and J. Sun. Faster R-CNN: Towards real-time object detection with region proposal networks. In *Advances in Neural Information Processing Systems*, pages 91–99, 2015.

[26] O. Ronneberger, P. Fischer, and T. Brox. U-Net: Convolutional Networks for Biomedical Image Segmentation. In *Medical Image Computing and Computer-Assisted Intervention (MICCAI)*, pages 234–241, 2015.

[27] O. Russakovsky, J. Deng, H. Su, J. Krause, S. Satheesh, S. Ma, Z. Huang, A. Karpathy, A. Khosla, M. Bernstein, A. C. Berg, and L. Fei-Fei. ImageNet Large Scale Visual Recognition Challenge. *International Journal of Computer Vision (IJCV)*, 115(3):211–252, 2015.

[28] K. Simonyan and A. Zisserman. Very Deep Convolutional Networks for Large-Scale Image Recognition. *CoRR*, abs/1409.1, 2014.

[29] J. T. Springenberg, A. Dosovitskiy, T. Brox, and M. Riedmiller. Striving for Simplicity: The All Convolutional Net. *arXiv preprint arXiv:1412.6806*, 2014.

[30] N. Srivastava, G. E. Hinton, A. Krizhevsky, I. Sutskever, and R. Salakhutdinov. Dropout: A Simple Way to Prevent Neural Networks From Overfitting. *Journal of Machine Learning Research*, 15(1):1929–1958, 2014.

[31] S. Sudholt and G. A. Fink. PHOCNet: A deep convolutional neural network for word spotting in handwritten documents. In *Proc. Int. Conf. on Frontiers in Handwriting Recognition*, 2016.

[32] N. Zhang, M. Paluri, M. Ranzato, T. Darrell, and L. Bourdev. PANDA: Pose Aligned Networks for Deep Attribute Modeling. In *Computer Vision and Pattern Recognition*, pages 1637–1644, 2014.

[33] B. Zhou, A. Lapedriza, J. Xiao, A. Torralba, and A. Oliva. Learning Deep Features for Scene Recognition Using Places Database. In *Advances in neural information processing systems*, pages 487–495, 2014.